%% file: 000_main.tex
\documentclass[11pt,a4paper]{article}
\usepackage[hyperref]{emnlp2020}
\usepackage{times}
\usepackage{latexsym}
\usepackage{graphicx}
\usepackage{tabularx}
\usepackage{color, colortbl}
\usepackage{amsmath}
\usepackage{booktabs}

\usepackage{microtype}

\aclfinalcopy 

\newcolumntype{Y}{>{\centering\arraybackslash}X}

\definecolor{Gray}{gray}{0.95}

\title{
ConveRT: Efficient and Accurate \\ Conversational Representations from Transformers
\\
{
        \ttfamily \small \href{https://github.com/PolyAI-LDN/polyai-models}{github.com/PolyAI-LDN/polyai-models}
}
}
\author{
 Matthew Henderson,
 I{\~{n}}igo Casanueva,
 Nikola Mrk{\v{s}}i\'c, \\
 {\bf 
 Pei-Hao Su, Tsung-Hsien Wen \normalfont{and}
 \textbf{Ivan Vuli\'{c}}
 } \\
 \texttt{\small matt@poly-ai.com} \\
 PolyAI Limited,
 London, UK
}

\date{}

\begin{document}
\maketitle
\begin{abstract}
\input{00_abstract.tex}
\end{abstract}

\section{Introduction}
\label{s:intro}
\input{01_introduction.tex}

\section{Methodology}
\label{s:methodology}
\input{02_methodology.tex}

\section{Experimental Setup}
\label{s:exp}
\input{03_experimental.tex}

\section{Results and Discussion}
\label{s:results}
\input{04_results}

\section{Conclusion}
\label{s:conclusion}
\input{06_conclusion.tex}

\clearpage
\bibliographystyle{acl_natbib}
\bibliography{refs}

\end{document}

%% file: 00_abstract.tex
General-purpose pretrained sentence encoders such as BERT are not ideal for real-world conversational AI applications; they are computationally heavy, slow, and expensive to train. We propose \textbf{ConveRT} (\textbf{Conve}rsational \textbf{R}epresentations from \textbf{T}ransformers), a pretraining framework for conversational tasks satisfying all the following requirements: it is effective, affordable, and quick to train. We pretrain using a retrieval-based response selection task, effectively leveraging quantization and subword-level parameterization in the dual encoder to build a lightweight memory- and energy-efficient model. We show that ConveRT achieves state-of-the-art performance across widely established response selection tasks. We also demonstrate that the use of extended dialog history as context yields further performance gains. Finally, we show that pretrained representations from the proposed encoder can be transferred to the intent classification task, yielding strong results across three diverse data sets. {ConveRT} trains substantially faster than standard sentence encoders or previous state-of-the-art dual encoders. With its reduced size and superior performance, we believe this model promises wider portability and scalability for Conversational AI applications.

%% file: 01_introduction.tex
Dialog systems, also referred to as conversational systems or conversational agents, have found use in a wide range of applications. They assist users in accomplishing well-defined tasks such as finding and booking restaurants, hotels, and flights \cite{Hemphill:1990,Williams:2012b,ElAsri:2017sigdial}, with further use in tourist information \cite{Budzianowski:2018emnlp}, language learning \cite{Raux:2003,Chen:2017survey}, entertainment \cite{Fraser:2018iva}, and  healthcare \cite{Laranjo:2018,Fadhil:2019arxiv}. They are also key components of intelligent virtual assistants such as Siri, Alexa, Cortana, and Google Assistant.

Data-driven task-oriented dialog systems require domain-specific labelled data: annotations for intents, explicit dialog states, and mentioned entities \cite{Williams:2014sigdial,Wen:17,Wen:2017icml,Ramadan:2018acl,Liu:2018naacl,Zhao:2019naacl}. This makes the scaling and maintenance of such systems very challenging. Transfer learning on top of pretrained models \cite[\textit{inter alia}]{Devlin:2018arxiv,Liu:2019roberta} provides one avenue for reducing the amount of annotated data required to train models capable of generalization.

Pretrained models making use of language-model (LM) based learning objectives have become prevalent across the NLP research community. When it comes to dialog systems, \emph{response selection} provides a more suitable pretraining task for learning representations that can encapsulate conversational cues. Such models can be pretrained using large corpora of natural unlabelled \emph{conversational} data \cite{Henderson:2019acl,Mehri:2019acl}. Response selection is also directly applicable to retrieval-based dialog systems, a popular and elegant approach to framing dialog  \cite{Wu:2017acl,Weston:2018ws,Mazare:2018emnlp,Gunasekara:2019dstc7,Henderson:2019acl}.\footnote{Retrieval-based dialog is popular because posing dialog as response selection \cite{Gunasekara:2019dstc7} simplifies system design  \cite{Boussaha:2019arxiv}. Unlike modular or end-to-end task-oriented systems, retrieval-based ones do not rely on dedicated modules for language understanding, dialog management, and generation. They mitigate the requirements for explicit task-specific semantics hand-crafted by domain experts \cite{Henderson:14a,Mrksic:15,Mrksic:17}.
}




\vspace{1.4mm}
\noindent \textbf{Response Selection} is a task of selecting the most appropriate \textit{response} given the dialog history \cite{Wang:2013emnlp,AlRfou:2016arxiv,Yang:2018repl,Du:2018scai,Chaudhuri:2018conll}. This task is central to retrieval-based dialog systems, which typically encode the \textit{context} and a large collection of responses in a joint semantic space, and then retrieve the most relevant response by matching the query representation against the encodings of each candidate response. The key idea is to: \textbf{1)} make use of large unlabelled conversational datasets (such as Reddit conversational threads) to \textit{pretrain} a neural model on the general-purpose response selection task; and then \textbf{2)} \textit{fine-tune} this model, potentially with additional network layers, using much smaller amounts of task-specific data. 

Dual-encoder architectures pretrained on response selection have become increasingly popular in the dialog community \cite{Cer:2018arxiv,Humeau:2019arxiv,Henderson:2019acl}. In recent work, \newcite{Henderson:2019arxiv} show that standard pretraining LM-based architectures cannot match the performance of dual encoders when applied to dialog tasks such as response retrieval.

\vspace{1.4mm}
\noindent \textbf{Scalability and Portability.} 
A fundamental problem with pretrained models is their large number of parameters (see Table~\ref{tab:comparison} later): they are typically highly computationally expensive to both train and run \cite{Liu:2019roberta}. 
 Such high memory footprints and computational requirements hinder quick deployment as well as their wide portability, scalability, and research-oriented exploration. The need to make pretrained models more compact has been recognized recently, with a line of work focused on building more efficient pretraining and fine-tuning protocols \cite{Tang:2019arxiv,Sanh:2019arxiv}. The desired reductions have been achieved through techniques such as distillation \cite{Sanh:2019arxiv}, quantization-aware training \cite{Zafrir:2019arxiv}, weight pruning \cite{Michel:2019nips} or weight tying \cite{Lan:2019albert}. However, the primary focus so far has been on optimizing the LM-based models, such as BERT.
 
 \vspace{1.4mm}
\noindent \textbf{ConveRT.} 
This work introduces a {\em more compact pretrained response selection model} for dialog. ConveRT is only 59MB in size, making it significantly smaller than the previous state-of-the-art dual encoder (444MB). It is also more compact than other popular sentence encoders, as illustrated in Table~\ref{tab:comparison}. This notable reduction in size and training acceleration are achieved through combining 8-bit embedding quantization and quantization-aware training, subword-level parameterization, and pruned self-attention. Furthermore, the lightweight design allows us to reserve additional parameters to improve the expressiveness of the dual-encoder architecture; this leads to \textit{improved learning of conversational representations} that can be transferred to other dialog tasks \cite{Casanueva:2020ws,Bunk:2020arxiv}. 


\vspace{1.4mm}
\noindent \textbf{Multi-Context Modeling.} 
ConveRT moves beyond the limiting single-context assumption made by \newcite{Henderson:2019acl}, where only the immediate preceding context was used to look for a relevant response. We propose a multi-context dual-encoder model which combines the immediate context with previous dialog history in the response selection task. The multi-context ConveRT variant remains compact (73MB in total), while offering improved performance on a range of established response selection tasks. We report significant gains over the previous state-of-the-art on benchmarks such as Ubuntu DSTC7 \cite{Gunasekara:2019dstc7}, AmazonQA \cite{Wan:2016icdm} and Reddit response selection \cite{Henderson:2019arxiv}, both in single-context and multi-context scenarios. Moreover, we show that sentence encodings learned by the model can be transferred to other dialog tasks, reaching strong intent classification performance over three evaluation sets. Pretrained dual-encoder models, both single-context and multi-context ones, are shared as TensorFlow Hub modules at \href{https://github.com/PolyAI-LDN/polyai-models}{\ttfamily \small github.com/PolyAI-LDN/polyai-models}.\footnote{Finally, our more compact neural response selection architecture is well aligned with the recent socially-aware initiatives on reducing costs and improving fairness and inclusion in NLP research and practice \cite{Strubell:2019acl,Mirzadeh:2019arxiv,Schwartz:2019green}. Cheaper training (pretraining the proposed dual-encoder model on the entire Reddit costs only 85 USD) and quicker development cycles offer new opportunities for more researchers and practitioners to tap into the construction of neural task-based dialog systems. }



%% file: 02_methodology.tex
\noindent \textbf{Pretraining on Reddit Data.} 
We assume working with English throughout the paper. Simplifying the conversational learning task to response selection, we can relate target dialog tasks to general-domain conversational data such as Reddit~\cite{AlRfou:2016arxiv}. This allows us to fine-tune the parameters of the task-specific response selection model, starting from the general-domain response selection model pretrained on Reddit. Similar to \newcite{Henderson:2019acl}, we choose Reddit  for pretraining due to: \textbf{1)} its organic conversational structure; and \textbf{2)} its unmatched size, as the public repository of Reddit data comprises 727M \textit{(input, response)} pairs.\footnote{\textcolor{darkblue}{github.com/PolyAI-LDN/conversational-datasets}}


\subsection{More Compact Response Selection Model}
\label{ss:main}
\begin{figure}[!t]
\centering
\includegraphics[width=0.87\linewidth]{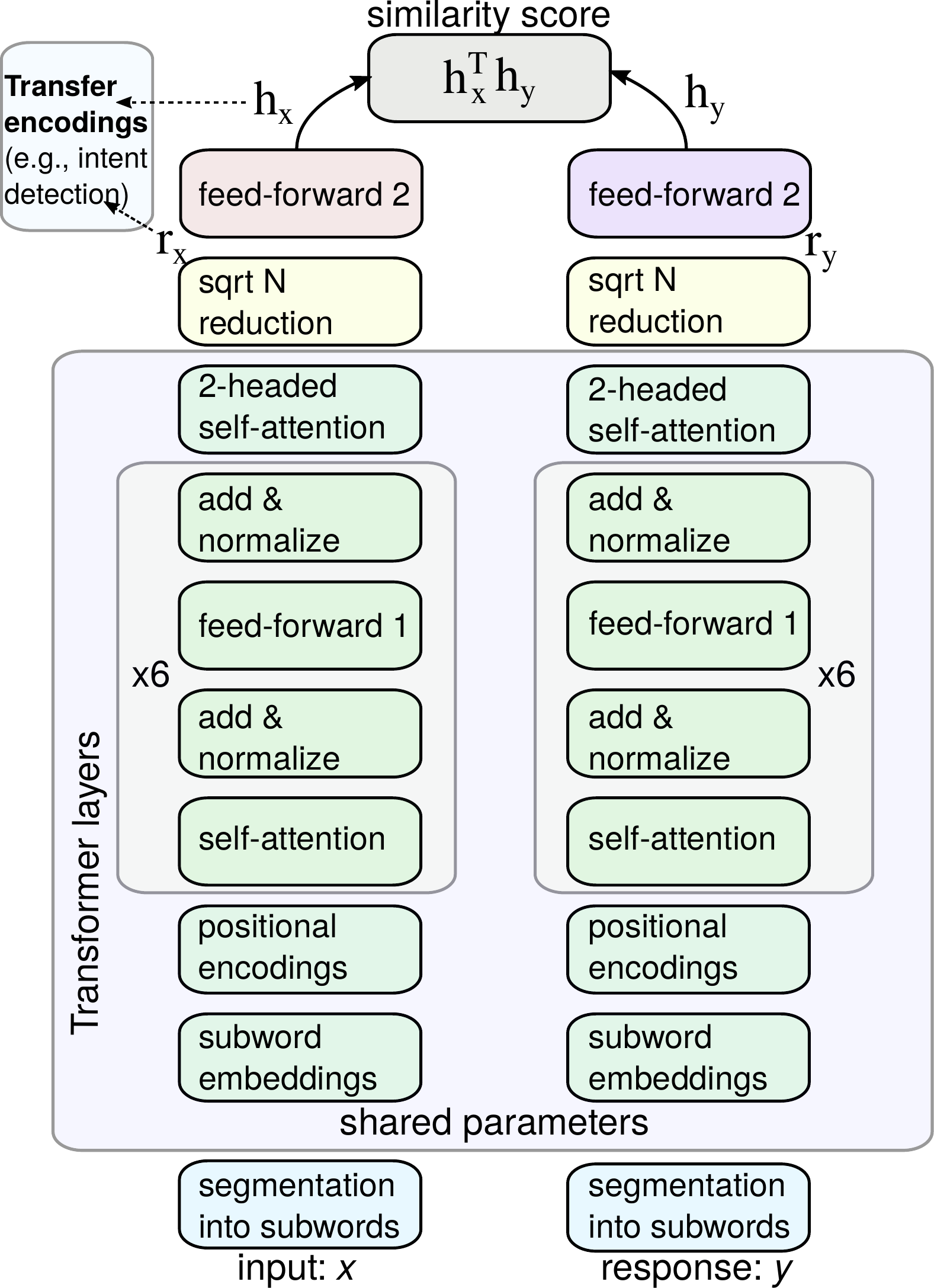}
\vspace{-2mm}
\caption{\textit{Single-context ConveRT} dual-encoder model architecture. Its multi-context extension is illustrated in Figure~\ref{fig:multi-enc}. It is possible to \textit{transfer} learned encodings at different network layers (e.g., $r_x$ or the final $h_x$) to other tasks such as intent detection or value extraction (see \S\ref{s:results}).
Note that the model uses two different feed-forward network (FFN) layers: 1) \textit{feed-forward 1} is the standard FFN layer also used by \newcite{Vaswani:2017nips}, and 2) \textit{feed-forward 2} contains 3 fully-connected non-linear feed-forward layers followed by a linear layer which maps to the final encodings $h_x$ and $h_y$ (note that the two \textit{feed-forward 2} networks do not share parameters, while the \textit{feed-forward 1} parameters are shared).
}
\vspace{-2.5mm}
\label{fig:main-enc}
\end{figure}

We propose \textbf{ConveRT} -- \textbf{C}onversational \textbf{R}epresentations from \textbf{T}ransformers -- 
a \textit{compact dual-encoder pretraining architecture}, leveraging subword representations, transformer-style blocks, and quantization, as illustrated in Figure~\ref{fig:main-enc}. ConveRT satisfies all the following requirements: it is effective, affordable, and quick to train.


\vspace{1.3mm}
\noindent \textbf{Input and Response Representation.} 
Prior to training, we obtain a vocabulary of subwords $V$ shared by the input side and the response side: we randomly sample and lowercase 10M sentences from Reddit, and then iteratively run any subword tokenization algorithm.\footnote{In the actual implementation, we use the same subword tokenization as \newcite{vaswani2018}. We run it for 4 iterations and retain only subwords occurring at least 250 times, containing no more than 20 UTF8 characters, also disallowing more than 4 consecutive digits.} The final vocabulary $V$ contains 31,476 subword tokens. During training and inference, if we encounter an OOV character it is treated as a subword token, where its ID is computed using a hash function, and it gets assigned to one of 1,000 additional ``buckets'' reserved for the OOVs. We therefore reserve parameters (i.e., embeddings) for the 31,476 subwords from $V$ and for the additional 1,000 OOV-related buckets. At training and inference, after the initial word-level tokenization on UTF8 punctuation and word boundaries, input text $x$ is split into subwords following a simple left-to-right greedy prefix matching \cite{vaswani2018}. We tokenize all responses $y$ during training in exactly the same manner.

\vspace{1.3mm}
\noindent \textbf{Input and Response Encoder Networks.}
The subword embeddings then go through a series of transformations on both the input and the response side. The transformations are based on the standard Transformer architecture \cite{Vaswani:2017nips}. Before going through the self-attention blocks, we add positional encodings to the subword embedding inputs. Previous work (e.g., BERT and related models) \cite[\textit{inter alia}]{Devlin:2018arxiv,Lan:2019albert} learns a fixed number of positional encodings, one for each position in the sequence, allowing the model to represent a fixed number of positions. Instead, we learn two positional encoding matrices of different sizes- $M^1$ of dimensionality [47, 512] and $M^2$ of dimensionality [11, 512]. An embedding at position $i$ is added to: $M^1_{i \mod 47} + M^2_{i \mod 11}$.\footnote{Note that since 47 and 11 are coprime, this gives $47\cdot11=517$ different possible positional encodings. Similar to the original (non-learned) positional encodings from \newcite{Vaswani:2017nips}, the rationale behind this choice of positional encoding is to allow the model to generalize to unseen sequence lengths.}


The next layers closely follow the original Transformer architecture with some notable differences. First, we set maximum relative attention \cite{Shaw:2018naacl} in the six layers to the following respective values: [3, 5, 48, 48, 48, 48].\footnote{We zero out in training and inference the attention scores for pairs of words if they are further apart than the set maximum relative attention values.} This also helps the architecture to generalize to long sequences and distant dependencies: earlier layers are forced to group together meanings at the phrase level before later layers model larger patterns. We use single-headed attention throughout the network.\footnote{Multi-headed attention requires running computations on 4-tensors: [batch, time, head, embedding], while for single-headed attention, this reduces to 3-tensors, and effectively speeds up training without hurting performance.}

Before going into a softmax, we add a bias to the attention scores that depends only on the relative positions: $\alpha_{ij} \rightarrow \alpha_{ij} + B_{n-i+j}$ where $B$ is a learned bias vector. This helps the model understand relative positions, but is much more computationally efficient than computing full relative positional encodings \cite{Shaw:2018naacl}. Again, it also helps the model generalize to longer sequences.

Six Transformer blocks use a 64-dim projection for computing attention weights, a 2,048-dim kernel (\textit{feed-forward 1} in Figure~\ref{fig:main-enc}), and 512-dim embeddings. Note that all Transformer layers use parameters that are fully shared between the input side and the response side. As in the Universal Sentence Encoder (\textsc{use}) \cite{Cer:2018arxiv}, we use square-root-of-N reduction to convert the embedding sequences to fixed-dimensional vectors. Two self-attention heads each compute weights for a weighted sum, which is scaled by the square root of the sequence length; the length is computed as the number of constituent subwords.\footnote{In fact, rather than computing the self-attended sequence, then reducing it, we reduce the attention weights accordingly, and then directly apply them via matrix multiplication to the input sequence to get the final reduced representation, that is, we \emph{fuse} these two operations. This is more computationally efficient, avoiding another 3-tensor multiplication.} The outputs of the reduction layer, labelled $r_x$ and $r_y$ in Figure~\ref{fig:main-enc}, are 1,024-dimensional vectors that are fed to the two ``side-specific'' (i.e., they do not share parameters) feed-forward networks.

In other words, the vectors $r_x$ and $r_y$ go through a series of $N_{f}$ $l$-dim feed-forward hidden layers ($N_{f}=3$; $l=1,024$) with skip connections, layer normalization, and orthogonal initialization. The activation function used in these networks and throughout the architecture is the fast GeLU approximation \cite{hendrycks2016gelu}: $GeLU(x) = x \sigma(1.702 x)$. The final layer is linear and maps the text into the final L2-normalized 512-dim representation: $h_x$ for the input text, and $h_y$ for the corresponding response text (Figure~\ref{fig:main-enc}).

\vspace{1.4mm}
\noindent \textbf{Input-Response Interaction.}
The relevance of each response to the given input is then quantified by the score $S(x,y)$, computed as cosine similarity with annealing between the encodings $h_x$ and $h_y$. It starts at 1 and ends at $\sqrt{d}$, linearly increasing over the first 10K training batches. Training proceeds in batches of $K$ \textit{(input, response)} pairs $(x_1,y_1), \ldots, (x_K,y_K)$. The aim of the objective is to distinguish between the true relevant response ($y_i$) and irrelevant responses (i.e., negative samples) $y_j, j\neq i$ for each input sentence $x_i$. The training objective for a single batch of $K$ pairs is as follows: $J = \sum_{i=1}^K S(x_i,y_i) - \sum_{i=1}^K \log \sum_{j=1}^{K} e^{S(x_i,y_j)}$. The goal is to maximize the score of positive training pairs $(x_i, y_i)$ and minimize the score of pairing each input $x_i$ with $K'$ negative examples, which are responses that are not associated with the input $x_i$: for simplicity, all other $K-1$ from the current batch are used as negative examples.


\vspace{1.3mm}
\noindent \textbf{Quantization.}
Very recent work has shown that large models of language can be made more compact by applying quantization techniques \cite{Han:2016iclr}: e.g., quantized versions of Transformer-based machine translation systems \cite{Bhandare:2019arxiv} and BERT \cite{Shen:2019arxiv,Zhao:2019arxiv,Zafrir:2019arxiv} are now available. In this work, we focus on enabling quantization-aware conversational pretraining on the response selection task. We show that the dual-encoder ConveRT model from Figure~\ref{fig:main-enc} can be also be trained in a quantization-aware manner. Rather than the standard 32-bits per parameter, all embedding parameters are represented using only 8 bits, and other network parameters with just 16 bits; they are trained in a quantization-aware manner by adapting the mixed precision training scheme from \newcite{Micikevicius:2018iclr}. It keeps shadow copies of each variable with 32bit Floating Point (FP32) precision, but uses FP16-cast versions in the computations and inference models. Some operations in the graph, however, require FP32 precision to be numerically stable: layer normalization, L2-normalization, and softmax in attention layers.

Again, following \newcite{Micikevicius:2018iclr}, the final loss is scaled by 128, and the updates to the shadow FP32 variables are scaled back by 1/128: this allows the gradient computations to stay well represented by FP16 (e.g., they will not get rounded to zero). The subword embeddings are stored using 8-bits per parameter, and the quantization range is adjusted dynamically through training. It is updated periodically to contain all of the embedding values that have so-far been learned, with room for growth above and below - 10\% of the range, or 0.01 - whichever is larger. Finally, quantization also allows doubling the batch size, which also has a favourable effect of increasing the number of negative examples in training. 


\begin{figure}[!t]
\centering
\includegraphics[width=0.79\linewidth]{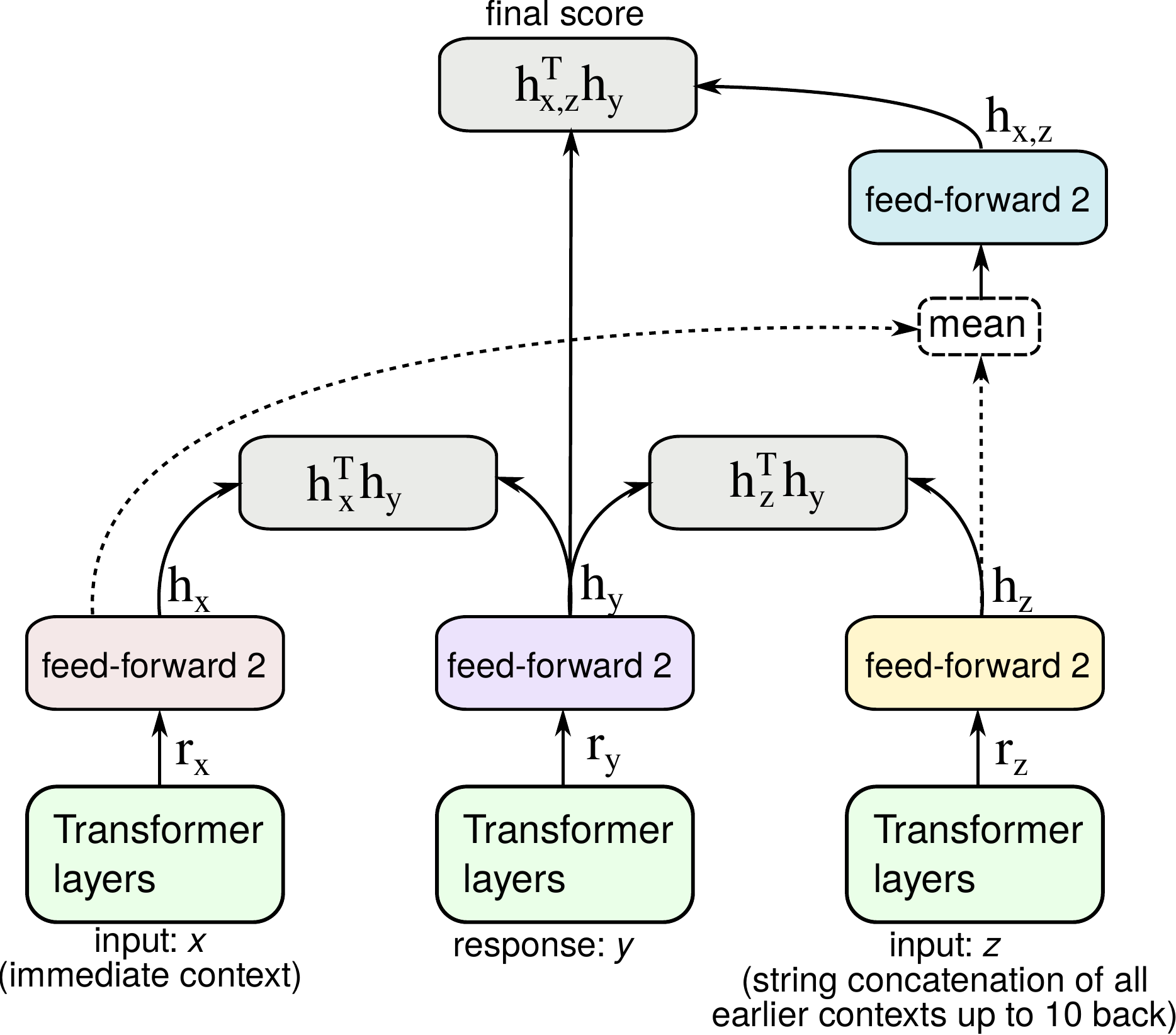}
\vspace{-1.5mm}
\caption{\textit{Multi-context ConveRT}. It models 1) the interaction between the immediate context and its accompanying response, 2) the interaction of the response with up to 10 earlier contexts from the conversation history, as well as 3) the interaction of the full context with the response. \textit{Transformer layers} refer to the standard Transformer architecture also used in the single-context encoder model in Figure~\ref{fig:main-enc}; the \textit{feed-forward 2} blocks are the same as with the single-context encoder architecture, see Figure~\ref{fig:main-enc}. The block \textit{mean} refers to simple averaging of two context encodings $h_x$ and $h_z$.}
\vspace{-1.5mm}
\label{fig:multi-enc}
\end{figure}


\vspace{1.4mm}
\noindent \textbf{Multi-Context ConveRT.}
\label{ss:contextual}
 Figure~\ref{fig:main-enc} depicts a single-context dual encoder architecture. Intuitively, the single-context assumption is limiting for modeling multi-turn conversations, where strong conversational cues can be found in earlier dialog history, and there has been a body of work on leveraging richer dialog history for response selection \cite{Chaudhuri:2018conll,Zhou:2018acl,Humeau:2019arxiv}. Taking a simple illustrative example: 


{\footnotesize
\vspace{1.4mm}
\indent \textbf{Student:} I'm very interested in representation learning. \\
\indent \textbf{Teacher:} Do you have any experience in PyTorch? \\
\indent \textbf{Student:} Not really. \\
\indent \textbf{Teacher:} And what about TensorFlow?}%
\vspace{1.4mm}

\noindent Selecting the last Teacher's response would be very difficult given only the immediate preceding context. However, the task becomes easier when taking into account the entire context of the conversation. We thus construct a \textit{multi-context dual-encoder model} by using up to 10 more previous messages in a Reddit thread. The extra 10 contexts are concatenated from most recent to oldest, and treated as an extra feature in the network, as shown in Figure~\ref{fig:multi-enc}. Note that all context representations are still independent from the representation of a candidate response, so we can still do efficient response retrieval and training. The full training objective is a linear combination of three sub-objectives: 1) ranking responses given the immediate context (i.e., this is equal to the single-context model from \S\ref{ss:main}), 2) ranking responses given only the extra (non-immediate) contexts, and 3) ranking responses given the averaged representation of the immediate context and additional contexts.\footnote{Combining multiple objectives in a dual-encoder framework has also been done by \newcite{AlRfou:2016arxiv} and \newcite{Henderson:2017arxiv}. Note that more sophisticated solutions to fusing dialog  history are possible such as using attention over older contexts as done by \newcite{Vlasov:2019arxiv} on the much smaller MultiWOZ 2.1 dataset \cite{Eric:2019arxiv}, but we have opted for simple concatenation as an efficient solution for training on the large Reddit data. The multiple objectives result in quicker learning, and also give useful diagnostic probes into the performance of each feature throughout training.}


%% file: 03_experimental.tex

\noindent \textbf{Training Data and Setup.} 
We base all our (pre)training on the large Reddit conversational corpus \cite{Henderson:2019arxiv} derived from 3.7B Reddit comments: it comprises 727M \textit{(input, response)} pairs for single-context modeling -- 654M pairs are reserved for training, the rest is used for testing. We truncate sequences to 60 subwords, embedding size is set to 512 for all subword embeddings and bucket embeddings, and the final encodings $h_x$, $h_y$, $h_z$, and $h_{x,z}$ are all 512-dimensional. The hidden layer size of \textit{feed forward 2} networks is set to 1,024 (with $N_f=3$ hidden layers used).

We train using ADADELTA with $\rho = 0.9$ \cite{Zeiler:2012ada}, batch size of 512, and a learning rate of 1.0 annealed to 0.001 with cosine decay over training. L2-regularization of $10^{-5}$ is used, subword embedding gradients are clipped to 1.0, and label smoothing of 0.2 is applied.\footnote{The label smoothing technique \cite{Szegedy:2016cvpr} reduces overfitting by preventing a network to assign full probability to the correct training example \cite{Pereyra:2017arxiv}. It means that each positive example in each batch is assigned the probability of 0.8, while the remaining probability mass is evenly redistributed across in-batch negative examples.} 


We pretrain the model on Reddit on 12 GPU nodes with one Tesla K80 each for 18 hours; this is typically sufficient to reach convergence. The total pretraining cost is roughly \$85 on Google Cloud Platform. This pretraining regime is orders of magnitude cheaper and more efficient than the prevalent pretrained NLP models such as BERT, GPT-2, XLNet, and RoBERTa \cite{Strubell:2019acl}.


\vspace{1.3mm}
\noindent \textbf{Baselines.} 
We report results on the response selection tasks and compare against the standard set of baselines \cite{Henderson:2019arxiv}. First, we compare to a simple keyword matching baseline based on \textsc{tf-idf} query-response scoring \cite{Manning:2008ir}, and then with a representative sample of publicly available neural encoders that embed inputs and responses into a vector space relying on various pretraining objectives: (1) The larger variant of Universal Sentence Encoder \cite{Cer:2018arxiv} (\textsc{use-large}); (2) The large variant of BERT \cite{Devlin:2018arxiv} (\textsc{bert-large}). We also compare to two recent dual-encoder architectures: (3) \textsc{use-qa} is a dual question-answer encoder version of the USE (large) model \cite{Chidambaram:2019repl}.\footnote{Note that \textsc{use-qa} encodes inputs/contexts and responses using separate sub-networks, while ConveRT (Figure~\ref{fig:main-enc}) relies on full parameter sharing in the Transformer layers.} (4) \textsc{polyai-dual} is the best-performing dual-encoder model from \newcite{Henderson:2019acl} pretrained on Reddit response selection. For baseline models 1-3, we report the results with the \textsc{map} response selection variant \cite{Henderson:2019arxiv}: it showed much stronger performance than a simpler similarity-based variant which directly ranks responses according to their cosine similarity with the context vector. \textsc{map} learns to (linearly) map the response vectors to the input vector space.


\vspace{1.3mm}
\noindent \textbf{Response Selection: Evaluation Tasks.}
We report response selection performance on Reddit test set \cite{Henderson:2019arxiv} with both single-context and multi-context ConveRT variants. For multi-context ConveRT, the averaged representation of (immediate and previous) context is used in evaluation. The models are applied directly on the Reddit test data without any further fine-tuning. We also evaluate on two other well-known response selection problems in different domains. (1) \textsc{amazonQA} \cite{Wan:2016icdm} is an e-commerce data set which contains information about Amazon products in the form of question-answer pairs:out of 3.6M (single-context) QA pairs, 300K pairs are reserved for testing. (2) \textsc{dstc7-ubuntu} is based on the Ubuntu v2 corpus \cite{Lowe:2017dd}: it contains 1M+ conversations in a highly technical domain (i.e., Ubuntu technical support). \textsc{dstc7-ubuntu} uses 100K conversations for training, 10K for validation, and 5K conversations are used for testing \cite{Gunasekara:2019dstc7}.


For \textsc{dstc7-ubuntu} we fine-tune for 60K training steps: it takes around 2h on 12 GPU workers. The learning rate starts at 0.1, and is annealed to 0.0001 using cosine decay over training. We use a batch size of 256, and dropout of 0.2 after the embedding and self-attention layers. We use the same fine-tuning regime for \textsc{amazonQA}. For \textsc{dstc7-ubuntu}, extra contexts are prepended with numerical strings 0--9 to help the model identify their position. We also release the fine-tuned models. 

We evaluate with a standard IR-inspired evaluation measure: \textit{Recall@k}, used in prior work on retrieval-based dialog \cite{Chaudhuri:2018conll,Henderson:2019acl,Gunasekara:2019dstc7}. Given a set of $N$ responses to the given input, where only one response is relevant, it indicates whether the relevant response occurs in the top $k$ ranked candidates. We denote this measure as $\mathbf{R}_{N}@k$, and set $N=100; k=1$: $\mathbf{R}_{100}@1$. 


\vspace{1.3mm}
\noindent \textbf{Intent Classification: Task, Data, Setup.} 
Pretrained sentence encoders have become particularly popular due to the success of training models
for downstream tasks on top of their learned representations, greatly improving the results compared
to training from scratch, especially in low-data regimes (see Table~\ref{tab:intent-data}). Therefore, we also probe the usefulness of ConveRT encodings for transfer learning in the intent classification task: the model must classify the user's utterance into one of several predefined classes, that is, \textit{intents} (e.g., within e-banking intents can be \textit{card lost} or \textit{replace card}). We use three internal intent classification datasets from three diverse domains, see Table~\ref{tab:intent-data}, divided into train, dev and test sets using a 80/10/10 split.

We use the pretrained ConveRT encodings $r_x$ on the input side (see Figure~\ref{fig:main-enc}) as input to an intent classification model. We also experimented with later $h_x$ encodings on the input side, but stronger results were observed with $r_x$. We train a 2-layer feed-forward net with dropout on top of $r_x$. SGD with a batch size of 32 is used, with early stopping after 5 epochs without improvement on the validation set. Layer sizes, dropout rate and learning rate are selected through grid search. We compare against two other standard sentence encoders again: \textsc{use-large} and \textsc{bert-large}. For ConveRT and \textsc{use-large} we keep the encoders fixed and train the classifier layers on top of the sentence encodings. For \textsc{bert-large}, we train on top of the \textsc{CLS} token and we fine-tune all its parameters.

\begin{table}[!t]
	\centering
    \def\arraystretch{0.85}
    {\small
	\begin{tabularx}{\linewidth}{l YY}
	    \toprule
	    {} & {\# intents} & {\# examples} \\
	    \cmidrule(lr){2-3}
	    {Banking (customer service)} & {77} & {14.6K} \\
	    {Shopping (online shopping)} & {10} & {13.8K} \\
	    {Company FAQ} & {110} & {3.3K}\\
		\bottomrule       
	\end{tabularx}}%
    \vspace{-2mm}
     \caption{Intent classification data sets.} \label{tab:intent-data}
     \vspace{-2.5mm}
\end{table}

%% file: 04_results.tex
\begin{table*}[!t]
	\centering
    \def\arraystretch{0.87}
    {\footnotesize
	\begin{tabularx}{\textwidth}{l YYYc}
	    \toprule
	    {} & \textbf{Embedding} & \textbf{Network} & \textbf{Total} & \textbf{Size after} \\
	    {} & \textbf{parameters} & \textbf{parameters} & \textbf{size} & \textbf{quantization} \\
	    \cmidrule(lr){2-5}
	    {USE \cite{Cer:2018arxiv}} & {256 M} & {2 M} & {1033 MB} & {261 MB} * \\
	    {BERT-BASE \cite{Devlin:2018arxiv}} & {23 M} & {86 M} & {438 MB} & {196 MB} */ 110 MB **\\
	    {BERT-LARGE \cite{Devlin:2018arxiv}} & {31 M} & {304 M} & {1341 MB} & {639 MB} */ 336 MB **\\
	    {GPT-2 \cite{radford2019language}} & {80 M} & {1462 M} & {6168 MB} & {3004 MB} * \\
	    {POLYAI-DUAL \cite{Henderson:2019acl}} & {104 M} & {7 M} & {444 MB} & {118 MB} \\
	    \cmidrule(lr){2-5}
	    \rowcolor{Gray}
	    {ConveRT (this work)} & {16 M} & {13 M} & {116 MB} & {\bf 59 MB} \\
		\bottomrule       
	\end{tabularx}}%
    \vspace{-1.5mm}
     \caption{Comparison of the proposed compact dual-encoder architecture for response selection to existing public standard sentence embedding models. (*) The size after quantization assumes embeddings can be quantized to 8 bits and network parameters to 16 bits,
which has not been verified for the public models. (**) Best-case model size estimates of the BERT model after full 8-bit quantization based on the work of \newcite{Zafrir:2019arxiv}.} \label{tab:comparison}
     \vspace{-2.5mm}
\end{table*}

\noindent \textbf{Model Size, Training Time, Cost.}
Table~\ref{tab:comparison} lists encoders from prior work along with their model size, and estimated model size after quantization. The reported numbers indicate the gains achieved through subword-level parameterization and quantization of ConveRT. Besides reduced training costs, ConveRT offers a reduced memory footprint and quicker training. We pretrain all our models for 18 hours only (on 12 16GB T4 GPUs), while a model compression technique DistilBERT \cite{Sanh:2019arxiv} (i.e., it reports $\approx 40\%$ relative reduction of the original BERT) trains on 8 16GB V100 GPUs for 90 hours, and larger models like RoBERTa require 1 full day of training on 1024 32GB V100 GPUs. The achieved size reduction and quick training also allow for quicker development and insightful ablation studies (see later in Table~\ref{tab:ablation}), and using quantization also improves training efficiency in terms of examples per second.

\begin{table}[!t]
	\centering
    \def\arraystretch{0.87}
    {\small
	\begin{tabularx}{\linewidth}{l YY}
	    \toprule
	    {} & {Reddit} & {AmazonQA} \\
	    \cmidrule(lr){2-3}
	    \textsc{tf-idf} & {26.4} & {51.8} \\
	    \textsc{use-large-map} & {47.7} & {61.9} \\
	    \textsc{bert-large-map} & {24.0} & {44.1} \\
	    \textsc{use-qa-map} & {46.6} & {70.7} \\
	    \textsc{polyai-dual} & {61.3} & {71.3} \\
	    \cmidrule(lr){2-3}
	    \rowcolor{Gray}
	    {ConveRT (single-context)} & {\it 68.2} & {84.3}\\
	    \rowcolor{Gray}
	    {ConveRT (multi-context)} & {\bf 71.8} & {--}\\
		\bottomrule       
	\end{tabularx}}%
    \vspace{-1.5mm}
     \caption{$\mathbf{R}_{100}@1 \times 100\%$ scores on Reddit test set and \textsc{amazonQA}. \textsc{polyai-dual} and {ConveRT} networks are fine-tuned on the training portion of \textsc{amazonQA}. Note that \textsc{amazonQA} by design supports only single-context response selection.} \label{tab:reddit}
     \vspace{-1mm}
\end{table}


\begin{table}[!t]
	\centering
    \def\arraystretch{0.87}
    {\small
	\begin{tabularx}{1.0\linewidth}{l Y}
	    \toprule
	    {Model Configuration} & {} \\ 
	    \cmidrule(lr){1-1}
	    {ConveRT} & {68.2} \\ 
	    \cmidrule(lr){1-1}
        {A: Multi-headed attention (8 64-dim heads)} & {68.5} \\ 
        {B: No relative position bias} & {67.8} \\ 
        {C: Without gradually increasing max attention span} & {67.7} \\ 
        {D: Only 1 OOV bucket} & {68.0} \\ 
        {E: 1-headed (instead of 2-headed) reduction} & {67.7} \\ 
        {F: No skip connections in \textit{feed forward 2}} & {67.8} \\ 
        {D + E + F} & {66.7} \\ 
        {B + C + D + E + F} & {66.6} \\ 
		\bottomrule       
	\end{tabularx}}%
    \vspace{-1.5mm}
     \caption{An ablation study illustrating the importance of different components in ConveRT: single-context response selection on Reddit ($\mathbf{R}_{100}@1$). Each experiment has been run for 966K steps (batch size 512).} \label{tab:ablation}
     \vspace{-1.5mm}
\end{table}

\begin{table}[!t]
	\centering
    \def\arraystretch{0.87}
    {\small
	\begin{tabularx}{\linewidth}{l YY}
	    \toprule
	    {} & {$\mathbf{R}_{100}@1$} & {MRR} \\
	    \cmidrule(lr){2-3}
	    {Best DSTC7 System} & {64.5} & {73.5} \\
	    {GPT*} & {48.9} & {59.5} \\
	    {BERT*} & {53.0} & {63.2} \\
	    {Bi-encoder \cite{Humeau:2019arxiv}} & {70.9} & {78.1} \\
	    \cmidrule(lr){2-3}
	    \rowcolor{Gray}
	    {ConveRT (single-context)} & {38.2} & {49.2}\\
	    \rowcolor{Gray}
	    {ConveRT (multi-context)} & {\bf 71.2} & {\bf 78.8}\\
		\bottomrule       
	\end{tabularx}}%
    \vspace{-1.5mm}
     \caption{Results on \textsc{dstc7-ubuntu}. (*) Scores for GPT and BERT taken from \newcite{Vig:2019dstc}.} \label{tab:ubuntu1}
     \vspace{-1.5mm}
\end{table}

\vspace{1.3mm}
\noindent \textbf{Response Selection on Reddit.}
The results are summarized in Table~\ref{tab:reddit}. Even single-context ConveRT achieves peak performance in the task, with substantial gains over the previous best reported score of \newcite{Henderson:2019acl}. It also substantially outperforms all the other models which were not pretrained directly on the response selection task, but on a standard LM task instead. The strongest baselines, however, are two dual-encoder architectures (i.e., \textsc{use-large}, \textsc{use-qa} and \textsc{polyai-dual}); this illustrates the importance of explicitly distinguishing between inputs/contexts and responses when modeling response selection. 

Table~\ref{tab:reddit} also shows the importance of leveraging additional contexts (see Figure~\ref{fig:multi-enc}). Multi-context ConveRT achieves a state-of-the-art Reddit response selection score of \textbf{71.8\%} We observe similar benefits in other reported response selection tasks. We also note the results of 1) using only the sub-network that models the interaction between the immediate context and the response (i.e., the $h_{x}^{T} h_{y}$ interaction), and 2) artificially replacing the concatenated extra contexts $z$ with an empty string. The respective scores are 65.7\% and 65.6\%. This suggests that multi-context ConveRT is also applicable to single-context scenarios when no extra contexts are provided for the target task. 

\vspace{1.3mm}
\noindent \textbf{Ablation Study.} 
The efficient training regime also allows us to perform a variety of diagnostic experiments and ablations. We report results with variants of single-context ConveRT in Table~\ref{tab:ablation}. They indicate that replacing single-headed with multi-headed attention leads to slight improvements, but this comes at a cost of slower (and consequently - more expensive) training. Using 1 instead of 1,000 OOV buckets leads only to a modest decrease in performance. Most importantly, the ablation study indicates that the final performance actually comes from the synergistic effect of applying a variety of components and technical design choices such as skip connections, 2-headed reductions, relative position biases, etc. While removing only one component at a time yields only modest performance losses, the results show that the loss adds up as we remove more components, and different components indeed contribute to the final score.\footnote{Furthermore, quick development and short training times also allow us to treat some of the component choices as hyper-parameter choices. It effectively means that such configuration choices can also be fine-tuned similar to any other hyper-parameter to optimize the final retrieval performance.}


\vspace{1.3mm}
\noindent \textbf{Other Response Selection Tasks.} 
The results on the \textsc{amazonQA} task are provided in Table~\ref{tab:reddit}. We see similar trends as with Reddit evaluation. Fine-tuned ConveRT reaches a new state-of-the-art score, and the strongest baselines are again dual-encoder networks. Fine-tuned \textsc{polyai-dual}, which was pretrained on exactly the same data, cannot match ConveRT's performance.\footnote{Interestingly, directly applying ConveRT to \textsc{amazonQA} without any fine-tuning also yields a reasonably high score of 67.0\%. Moreover, learning the mapping function between inputs and responses (again without any fine-tuning) for ConveRT the same way as is done for \textsc{use-qa-map} results in the score of 71.6\%, which outperforms \textsc{use-qa-map} (70.7\%). The gap to the fine-tuned model's performance, however, indicates the importance of in-domain fine-tuning.}

The results on \textsc{dstc7-ubuntu} are summarized in Table~\ref{tab:ubuntu1} First, they suggest very competitive performance of multi-context ConveRT model: it outperforms the best-scoring system from the official DSTC7 challenge \cite{Gunasekara:2019dstc7}. It is an encouraging finding, given that multi-context ConveRT relies on simple context concatenation without any additional attention mechanisms. We leave the investigation of such more sophisticated models to integrate additional contexts for future work. Multi-context ConveRT can also match or even surpass the performance of another dual-encoder architecture from \newcite{Humeau:2019arxiv}. Their dual encoder (i.e., \textit{bi-encoder}) is based on the BERT-base architecture \cite{Humeau:2019arxiv}: it relies on 12 Transformer blocks, 12 attention heads, and a hidden size dimensionality of 768 (while we use 512). Training with that model is roughly 5$\times$ slower, and the pretraining objective is more complex: they use the standard BERT pretraining objective plus next utterance classification. Moreover, their model is trained on 32 v100 GPUs for 14 days, which makes it roughly 50$\times$ more expensive than ConveRT.

\begin{table}[!t]
	\centering
    \def\arraystretch{0.85}
    {\small
	\begin{tabularx}{\linewidth}{l YYc}
	    \toprule
	    {} & {Banking} & {Shopping} & {Company FAQ} \\
	    \cmidrule(lr){2-4}
	    \textsc{use-large} & {92.2} & {94.0} & {62.4} \\
	    \textsc{bert-large} & {93.2} & {94.3} & {61.2} \\
	    \cmidrule(lr){2-4}
	    \rowcolor{Gray}
	    {ConveRT} & {92.7} & {94.5} & {64.3}\\
		\bottomrule       
	\end{tabularx}}%
    \vspace{-1.5mm}
     \caption{Intent classification results.} \label{tab:intent}
     \vspace{-2.5mm}
\end{table}


\vspace{1.3mm}
\noindent \textbf{Intent Classification.}
The results are summarized in Table~\ref{tab:intent}: we report the results of two strongest baselines for brevity. The scores show very competitive performance of ConveRT encodings $r_x$ transferred to another dialog task. They outperform \textsc{use-large} in all three tasks and \textsc{bert-large} in 2/3 tasks. Note that, besides quicker pretraining, intent classifiers based on ConveRT encodings train 40 times faster than \textsc{bert-large}-based ones, as only the classification layers are trained for ConveRT. In sum, these preliminary results suggest that ConveRT as a sentence encoder can be useful beyond the core response selection task. The usefulness of ConveRT-based sentence representations have been recently confirmed on other intent classification datasets \cite{Casanueva:2020ws}, with different intent classifiers \cite{Bunk:2020arxiv}, and in another dialog task: turn-based value extraction \cite{Coope:2020acl,Bunk:2020arxiv}. In future work, we plan to investigate other possible applications of transfer, especially for low-data setups. 

%% file: 06_conclusion.tex
We have introduced ConveRT, a new light-weight model of neural response selection for dialog, based on Transformer-backed dual-encoder networks, and have demonstrated its state-of-the-art performance on an array of response selection tasks and in transfer learning for intent classification tasks. In addition to offering \textit{more accurate} conversational pretraining models this work has also resulted in \textit{more compact} conversational pretraining. The quantized versions of ConveRT and multi-context ConveRT take up only 59 MB and 73 MB, respectively, and train for 18 hours with a training cost estimate of only 85 USD. In the hope that this work will motivate and guide further developments in the area of retrieval-based task-oriented dialog, we publicly release pretrained ConveRT models.

